%% file: main.tex
\documentclass{article}

\usepackage[preprint]{neurips_2026}

\usepackage[utf8]{inputenc} % allow utf-8 input
\usepackage[T1]{fontenc}    % use 8-bit T1 fonts
\usepackage{hyperref}       % hyperlinks
\usepackage{url}            % simple URL typesetting
\usepackage{booktabs}       % professional-quality tables
\usepackage{amsfonts}       % blackboard math symbols
\usepackage{nicefrac}       % compact symbols for 1/2, etc.
\usepackage{microtype}      % microtypography
\microtypesetup{expansion=false}
\usepackage{xcolor}         % colors
\usepackage[normalem]{ulem} % strikethrough for revision tracking
\usepackage{algorithm}      % algorithm floats
\usepackage{algpseudocode}  % pseudocode
\usepackage{graphicx}       % figures
\usepackage{eso-pic}        % page overlays
\usepackage{tikz}           % system diagrams
\usepackage{amsmath}
\usetikzlibrary{arrows.meta,positioning,calc,fit,backgrounds}

\makeatletter
\renewcommand{\@noticestring}{}
\makeatother

\definecolor{OmegaBlue}{HTML}{2E86EB}

\newcommand{\firstpagelogo}{%
  \AddToShipoutPictureFG*{%
    \AtPageUpperLeft{%
      \raisebox{-0.8in}{\hspace{1.35in}\includegraphics[width=1.35in]{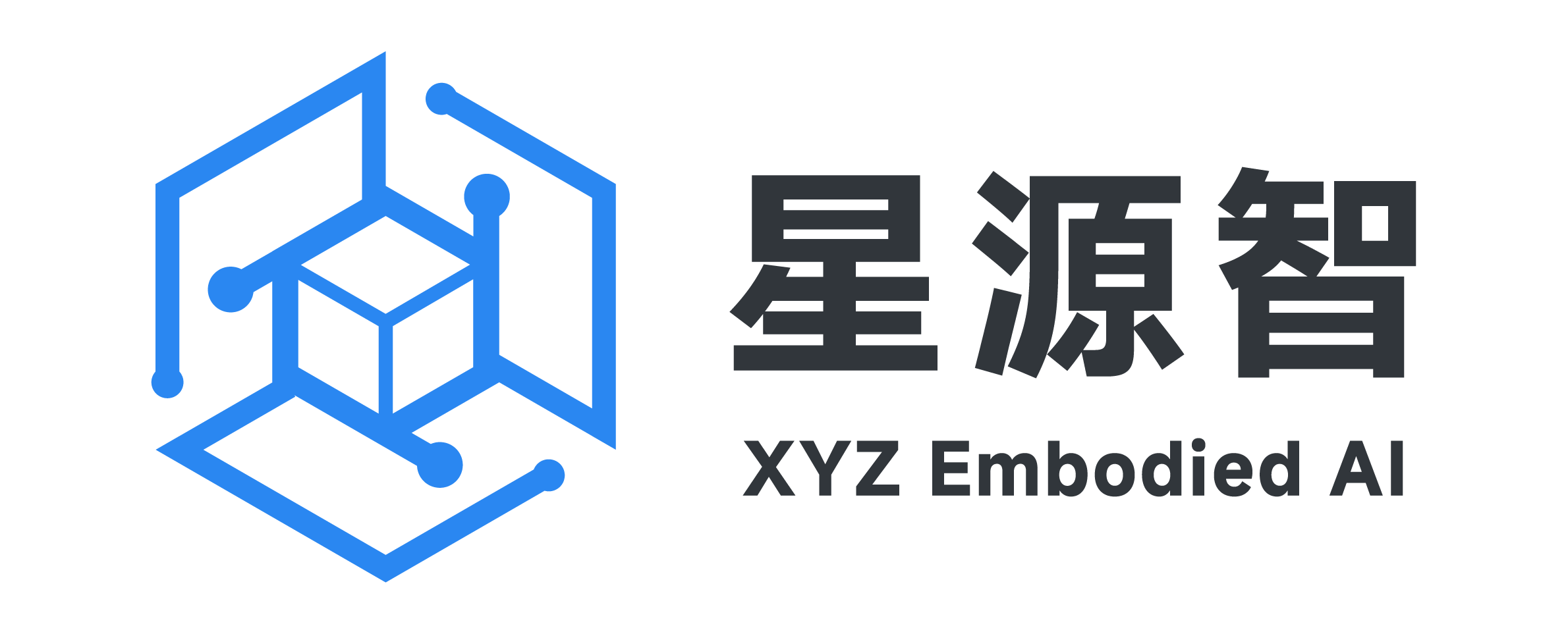}}%
    }%
  }%
}

\title{\textcolor{OmegaBlue}{$\omega$-EVA: Envision, Verify, and Act with Latent Interactive World Models}}

\author{Zhenguo Sun\thanks{Contributed equally.}\quad
Yu Sun\footnotemark[1] \thanks{Project leader.}\quad
Hande Huang\quad Alois Knoll}

\begin{document}

\firstpagelogo
\maketitle

\begin{abstract}
Embodied policies typically map current observations directly to actions, leaving candidate-action consequences implicit. World models provide predictive supervision, representations, or external simulation, but rarely let a policy inspect the imagined consequence of its own proposal before acting. We introduce $\omega$-EVA, a latent interactive world model that realizes an Envision--Verify--Act loop for embodied action generation. Its three-stage framework learns action-conditioned latent dynamics, trains a language-conditioned flow policy on dynamics-aware visual representations, and feeds the policy's proposal back through the world model. A tri-branch refiner jointly reasons over the current state, proposal-conditioned future, and proposed action to produce the final action chunk. Because consequence reasoning remains in latent feature space, $\omega$-EVA avoids generating future videos at inference. Evaluations across diverse single-arm, bimanual, long-horizon, and perturbed simulation settings show that the complete interaction pipeline consistently improves the proposal policy, while latent diagnostics indicate meaningful action-conditioned future structure. With approximately 1.2B parameters and no additional robot-data pretraining, $\omega$-EVA demonstrates a compact and competitive performance--scale--data trade-off, making the world model an active action-feedback module rather than a passive predictor.
Project page: \url{https://baai-humanoid.github.io/Omega-EVA/}. Notice: This project is not done yet, we are still working on ablation studies and real-world experiments. 
\end{abstract}

\input{chapters/1_Intro}
\input{chapters/2_Related_work}
\input{chapters/3_Method}
\input{chapters/4_Experiments}

\input{chapters/5_Conclusion.tex}

\bibliographystyle{plainnat}
\bibliography{references}

\appendix
\input{chapters/6_Appendix}

\end{document}

%% file: chapters/1_Intro.tex
\section{Introduction}

Embodied manipulation is inherently counterfactual. A robot should not only ask which action best matches the current observation and language instruction; it should also ask what would happen if that action were executed. This distinction becomes critical in dexterous manipulation, where small errors in reaching, grasping, or object alignment can compound into failure. Recent vision-language-action (VLA) models and generative visuomotor policies have made impressive progress on the first question. Large-scale robot transformers and VLA models show strong language-conditioned generalization across diverse manipulation tasks \citep{brohanRT1RoboticsTransformer2023,kimOpenVLAOpenSourceVisionLanguageAction2024,black2024pi_0,intelligence2025pi_}, while action-chunking and diffusion-style policies improve the modeling of continuous, multimodal control trajectories \citep{zhaoLearningFineGrainedBimanual2023,chiDiffusionPolicyVisuomotor2025,liuRDT1BDiffusionFoundation2025}. Yet most policies still follow a direct observation-to-action paradigm: given the present scene, decode an action chunk. The future consequence of that chunk remains implicit inside the policy parameters, rather than being exposed as something the policy can inspect before acting.

World models appear to offer the missing counterfactual interface by modeling how the environment evolves under condition \citep{worldmodelsurvey,hou2026worldsurvey}. Recent world-action models extend this idea to robotics by jointly learning visual dynamics and robot actions, showing that future prediction can inject useful physical structure into policy learning \citep{zhu2025uwm,li2025unified,ma2026dit4dit,bi2026motus,team2026motubrain}. However, existing approaches usually make the world model useful in one of three ways. Some use future video or latent prediction mainly as a training-time auxiliary objective or representation-learning signal, so explicit imagination is skipped during inference for efficiency \citep{li2025unified,ma2026dit4dit,yuan2026fastwam}. Others perform test-time video generation, goal-conditioned prediction, or rollout-based planning, which can provide richer consequence reasoning but is expensive for closed-loop robot control \citep{ye2026dreamzero,guo2025ctrlworld,zhou2025act2goal}. A third line uses predictive models as external guidance or conditioning signals for a generative policy \citep{an2026feedback}. These uses are valuable, but they rarely create an internal interaction loop in which a candidate action is tested against an action-conditioned world model and corrected before execution.

We introduce $\omega$-EVA, a latent interactive world model for embodied action generation. $\omega$-EVA is built around a simple but different paradigm: a policy should interact with its own imagined consequence before it acts. Instead of treating the world model as a passive auxiliary predictor or a standalone video simulator, $\omega$-EVA makes it an active verifier inside the action-generation step. The policy first proposes an action chunk, the world model envisions the latent future induced by that exact proposal, and a refiner updates the action using both the proposal and its imagined consequence. This Envision--Verify--Act loop turns future prediction into proposal-conditioned feedback: the robot does not merely predict what to do from the present, but checks what its intended action is likely to cause.

$\omega$-EVA realizes this paradigm with a three-stage training procedure. First, we pretrain an action-conditioned latent world model that predicts future visual features from current visual features and an action chunk, yielding both a future latent prediction and a dynamics-aware current representation. Second, we train a language-conditioned action generation policy on these world-model-aware current latents, so the initial action proposal already benefits from action-conditioned visual dynamics. Third, and most importantly, we freeze the pretrained world model and policy, feed the policy's own proposal back through the world model, obtain the imagined future latent caused by that proposal, and train a refiner over the current latent, imagined future latent, and proposed action. It therefore changes the role of the world model from ``learn a better representation'' to ``interact with a concrete action candidate and help correct it.''

At inference time, $\omega$-EVA follows the same Envision--Verify--Act loop. Given a visual observation and language instruction, the policy generates an initial action proposal. The frozen latent world model then predicts the future visual features that would result from executing this proposal. Finally, the refinement module verifies the proposal through the imagined consequence and outputs a refined action sequence for execution. Because imagination is performed in a compact feature space rather than through pixel-level video generation, $\omega$-EVA preserves test-time consequence reasoning while keeping the loop practical for closed-loop control. This design is especially relevant under visual perturbations, object-layout shifts, and contact uncertainty, where a proposal-conditioned consequence signal can help correct brittle action chunks before they reach the robot.

Our contributions are threefold. First, we propose latent interactive world modeling as a paradigm for embodied action generation, where a policy interacts with imagined consequences before execution. Second, we introduce an action-conditioned latent world model that predicts future visual features while providing dynamics-aware current representations for action generation. Third, we develop an Envision--Verify--Act refiner that turns stage 3 into proposal-conditioned consequence feedback, improving action robustness without requiring full future video generation.

%% file: chapters/2_Related_work.tex
\section{Related Work}

\subsection{Vision-Language-Action and Generative Visuomotor Policies}

Robot manipulation has increasingly shifted from task-specific visuomotor policies toward generalist models conditioned on vision and language. Large-scale robot transformers first demonstrated that data scale and task diversity can support broad real-world control \citep{brohanRT1RoboticsTransformer2023}. OpenVLA and subsequent vision-language-action (VLA) models extend this direction through open model development, flow-based action generation, and open-world generalization \citep{kimOpenVLAOpenSourceVisionLanguageAction2024,black2024pi_0,intelligence2025pi_}. Human-centric pretraining further transfers motion and interaction priors from large-scale human videos to dexterous and cross-embodiment robot learning \citep{luoBeingH0VisionLanguageActionPretraining2025,luoBeingH05ScalingHumanCentric2026}. Together, these works establish increasingly capable visual-language representations for direct robot control.

A complementary line focuses on the action distribution itself. Action Chunking Transformer predicts temporally extended action sequences \citep{zhaoLearningFineGrainedBimanual2023}, while diffusion-based policies capture multimodal continuous trajectories in visuomotor and bimanual manipulation \citep{chiDiffusionPolicyVisuomotor2025,liuRDT1BDiffusionFoundation2025}. Flow-based VLAs similarly generate action chunks by transporting noise toward a language- and observation-conditioned action distribution \citep{black2024pi_0}. These policy families provide strong proposal generators, but their standard interface remains observation and language in, action out. Although future consequences may be implicitly encoded in the learned parameters, the policy does not typically expose the visual consequence of its particular candidate action and use it as feedback before execution. $\omega$-EVA retains a generative action policy, but inserts an action-conditioned predictive interaction between proposing and finalizing the action.

\subsection{World-Action Models and Latent Predictive Learning}

World models learn useful structure by representing or predicting how the environment evolves \citep{worldmodelsurvey}. In embodied control, actions intervene directly on that evolution, motivating world-action models that jointly reason about observations, dynamics, and robot behavior \citep{hou2026worldsurvey,wang2026wamsurvey}. Existing systems use this predictive capability in several ways. Unified World Models and related video-action models couple future generation with action modeling for large-scale pretraining or policy learning \citep{zhu2025uwm,li2025unified,ma2026dit4dit}. Other systems treat world modeling as a scalable data engine or a unified control backbone \citep{team2025gigaworld0,team2026motubrain,li2026lingbotva}. Latent action world models and action-centered designs further seek compact and efficient interfaces between visual dynamics and control \citep{bi2026motus,ye2026gigaworldpolicy}. Collectively, these approaches show that future prediction can supply useful dynamics structure, supervision, and synthetic experience for robot policies.

Predictive representation learning offers an alternative to reconstructing full future videos. Video Prediction Policy learns future-oriented visual representations for generalist control \citep{hu2024vpp}, while JEPA style VLA incorporate latent future prediction into VLA learning \citep{sun2026vlajepa,miao2026jepavla}. More general joint-embedding architectures study stable latent prediction for visual understanding and planning \citep{maes2026leworldmodel,balestriero2025lejepa,assran2025v}. $\omega$-EVA shares their motivation for avoiding unnecessary pixel generation and likewise predicts future visual features. The distinction lies in the predictive interface: latent prediction is not used only to pretrain a representation or regularize a policy. At inference, $\omega$-EVA conditions the world model on the policy's own action proposal and exposes the resulting future latent to a separate action-refinement module.

\subsection{Test-Time Imagination and Consequence-Aware Action Refinement}

The foundational promise of world models is that imagined dynamics can inform decisions before they are executed \citep{ha2018world}. Dreamer and its successors operationalize this principle by learning behavior through latent rollouts in model-based reinforcement learning \citep{hafner2019dreamerv1,hafner2020dreamerv2,hafner2023dreamerv3}. In robot learning, recent work has adapted test-time imagination through several interfaces. Generative world-action models produce controllable rollouts or directly act as zero-shot policies \citep{ye2026dreamzero,guo2025ctrlworld}. Video models can also support visuomotor planning, generate policy targets, or translate predicted motion into robot actions \citep{kim2026cosmos,du2023unnipi,wen2024vidman,feng2025vidar,bharadhwaj2024gen2act,zhou2025act2goal}. These methods make predicted futures operational, but commonly rely on generated video trajectories, goals, or planning-oriented outputs.

Another direction asks whether explicit future generation is necessary at deployment. Fast-WAM studies efficient policies that avoid costly test-time imagination \citep{yuan2026fastwam}, while action-centered world-action models seek more efficient coupling between dynamics and actions \citep{ye2026gigaworldpolicy}. Feedback World Model instead uses predictive feedback to guide a diffusion policy during generation \citep{an2026feedback}. These approaches move world models closer to the policy loop, yet existing methods rarely combine three properties simultaneously: conditioning imagination on the policy's specific candidate action, feeding the imagined latent back within the same control decision, and jointly reasoning over the current state, imagined consequence, and original proposal to directly rewrite that action.

$\omega$-EVA closes this local interaction loop through proposal-conditioned latent feedback. Its policy first generates an action chunk; the frozen world model predicts the latent consequence of that exact proposal; and the refiner directly produces a new action from the present state, imagined future, and proposal. This is neither rollout-based planning nor reward evaluation, and it does not require decoding a future video. The world model instead serves as an internal action-feedback module, turning future prediction from an auxiliary learning signal or external simulator into an active participant in embodied action generation.

%% file: chapters/3_Method.tex
\section{Method}

\subsection{Problem Formulation}

We consider language-conditioned robot manipulation from visual observations. At environment step $t$, the robot receives a visual observation $o_t$ and a language instruction $l$, and must predict an action chunk $a_{t:t+H}\in\mathbb{R}^{H\times d_a}$, where $H$ is the action horizon and $d_a$ is the action dimension. Each training sample contains the current observation, instruction, and expert action chunk. To supervise action-conditioned dynamics, it additionally provides a future observation $o_{t+n}$ at a variable prediction step $n\in\{0,\ldots,H\}$, bounded by the remaining length of the episode.

A frozen visual encoder maps the current observation $o_t$ and future observation $o_{t+n}$ to patch-level features $I_c$ and $I_f^{(n)}$, respectively. Here, $I_f^{(n)}$ denotes the ground-truth future feature used to supervise latent dynamics prediction. A frozen text encoder maps the instruction $l$ to language tokens $T_e$. From $I_c$, the world model also learns a dynamics-aware current representation $c_t$ that captures visual information relevant to how the scene can evolve under robot actions.

A conventional visuomotor policy directly models the mapping $(o_t,l)\mapsto a_{t:t+H}$, leaving the consequence of the predicted action implicit. $\omega$-EVA instead introduces an action-conditioned latent consequence between action proposal and final action generation. Its inference problem is expressed by the composition
\[
\hat a^0_{t:t+H}
=
\pi_\phi(c_t,T_e),
\qquad
\hat I_f
=
W_\theta(I_c,\hat a^0_{t:t+H}),
\qquad
\hat a_{t:t+H}
=
R_\psi(c_t,\hat I_f,\hat a^0_{t:t+H}),
\]
where $\pi_\phi$ first produces an initial action proposal $\hat a^0_{t:t+H}$, the action-conditioned world model $W_\theta$ predicts its latent future consequence $\hat I_f$, and the refiner $R_\psi$ outputs the final action chunk $\hat a_{t:t+H}$ using the current representation, imagined future, and proposal itself. The learning objective is therefore not only to imitate expert actions, but to make the predicted consequence of a candidate action available as explicit feedback before that action is executed.

We next describe how $\omega$-EVA learns the latent dynamics model, proposal policy, and consequence-aware refiner through a three-stage framework.

\subsection{$\omega$-EVA Overview}

\begin{figure*}[htbp]
    \centering
    \includegraphics[width=\linewidth]{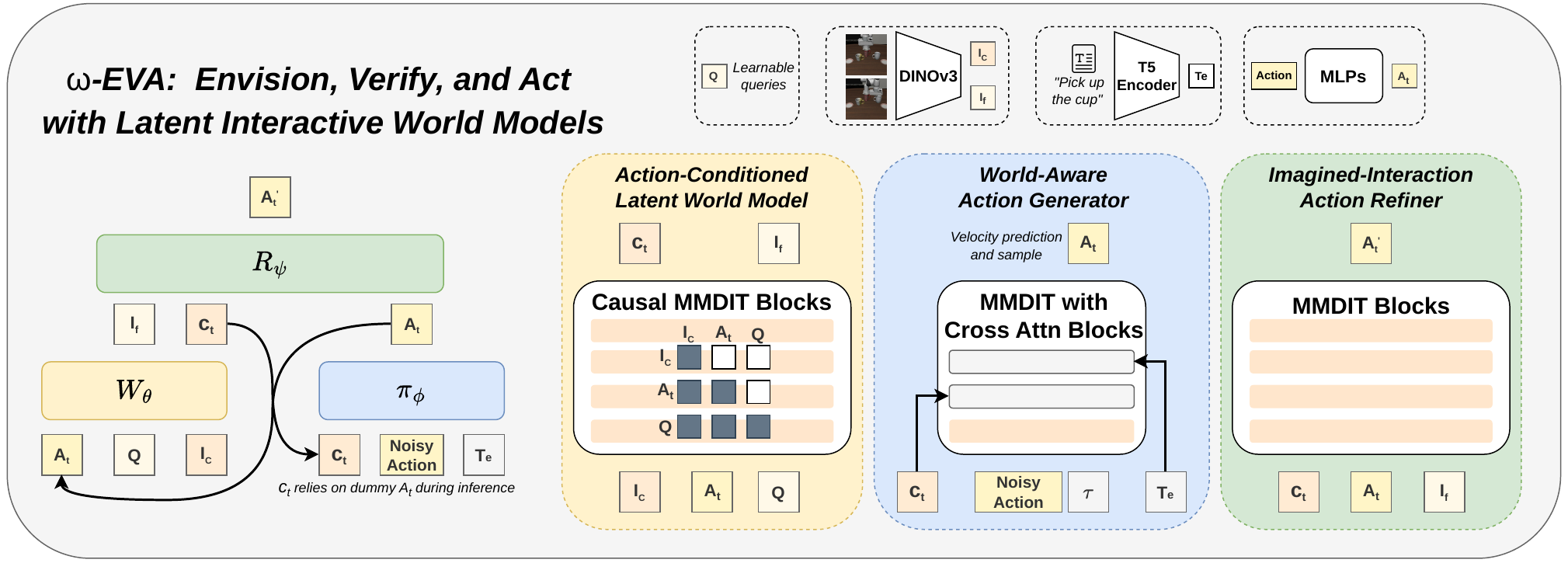}
    \caption{\textbf{Overview of $\omega$-EVA.} The left side shows the Envision--Verify--Act interaction loop: the world-aware action generator proposes an action, the action-conditioned latent world model envisions its consequence, and the imagined-interaction refiner corrects the proposal before execution. The right side details the three modules learned across Stages~1--3.}
    \label{fig:omega_eva_overview}
\end{figure*}

Figure~\ref{fig:omega_eva_overview} presents $\omega$-EVA as a closed interaction loop between action generation and latent consequence prediction. Given the current visual feature $I_c$ and language tokens $T_e$, the proposal policy $\pi_\phi$ first produces an initial action chunk $\hat a^0_{t:t+H}$ from the dynamics-aware current representation $c_t$. The world model $W_\theta$ then conditions on this specific proposal to envision its future consequence $\hat I_f$. Finally, the refiner $R_\psi$ jointly reasons over $c_t$, $\hat I_f$, and $\hat a^0_{t:t+H}$ to produce the refined action $\hat a_{t:t+H}$. The world model thus supports action generation in two complementary ways: it provides a dynamics-shaped representation of the observed state and an explicit consequence of the action currently under consideration.

$\omega$-EVA builds the capabilities required for this loop in three stages. Stage~1 learns the action-conditioned latent dynamics model $W_\theta$ by predicting variable-horizon future visual features. In addition to its future prediction, the model produces $c_t$, which preserves the current observation while encoding dynamics-relevant visual structure. Stage~2 learns the language-conditioned proposal policy $\pi_\phi$ on top of $c_t$ and $T_e$, yielding an initial action generator informed by the learned dynamics representation. Stage~3 freezes $W_\theta$ and $\pi_\phi$, feeds each generated proposal back into the world model, and trains $R_\psi$ to correct the proposal from its imagined latent consequence. Stages~1 and~2 therefore establish the predictive model and proposal policy, while Stage~3 connects them into the Envision--Verify--Act interaction loop.

Training Stage~3 uses the same proposal--imagination--refinement data flow as inference. At deployment, the refined action is executed without decoding $\hat I_f$ into pixels, and the full procedure is repeated after the next observation. This preserves explicit test-time consequence reasoning while keeping imagination inside a compact latent space. Algorithm~\ref{alg:omega_train} summarizes the optimization schedule; the following subsections detail the three learned components.

\begin{algorithm}[t]
\caption{Three-stage training of $\omega$-EVA}
\label{alg:omega_train}
\begin{algorithmic}[1]
\Require Training tuples $(I_c,I_f^{(n)},l,a_{t:t+H})$, with valid-action masks and $n$ bounded by the remaining episode length
\Ensure World model $W_\theta$ (including learnable future queries $Q$), flow policy $\pi_\phi$, and refiner $R_\psi$
\State \textbf{Stage 1: action-conditioned latent world model}
\For{each Stage 1 minibatch}
    \State Sample $n\in\{1,\ldots,\min(H,L_t)\}$ and construct $a_{t:t+H}^{(n)}$.
    \State Predict $(\hat I_f,c_t)=W_\theta(I_c,a_{t:t+H}^{(n)};Q)$.
    \State Update $\theta$, including $Q$, by minimizing $\mathcal{L}_{wm}=\|\hat I_f-I_f^{(n)}\|_1$.
\EndFor
\State \textbf{Stage 2: world-aware action generator}
\For{each Stage 2 minibatch}
    \State Encode instruction $l$ into language tokens $T_e$.
    \State Obtain $c_t$ from $W_\theta$ using the full expert action chunk.
    \State Sample $\tau$ and $\epsilon$, form $x_\tau=(1-\tau)a+\tau\epsilon$.
    \State Predict $v_\phi(x_\tau,c_t,T_e,\tau)$ and compute $\mathcal{L}_{fm}$.
    \State Update $\phi$ and, when unfrozen, $\theta$ (including $Q$) using $\mathcal{L}_{stage2}$.
\EndFor
\State \textbf{Stage 3: imagined-interaction action refiner}
\State Freeze $W_\theta$ (including $Q$) and $\pi_\phi$.
\For{each Stage 3 minibatch}
    \State Extract $c_t$ from $W_\theta(I_c,a_{\mathrm{dummy}};Q)$ and encode $l$ into $T_e$.
    \State Generate proposal $\hat a^0_{t:t+H}=\pi_\phi(c_t,T_e)$ by flow integration.
    \State Envision $(\hat I_f,c_t)=W_\theta(I_c,\hat a^0_{t:t+H};Q)$ without gradients.
    \State Refine action $\hat a_{t:t+H}=R_\psi(c_t,\hat I_f,\hat a^0_{t:t+H})$.
    \State Update $R_\psi$ by minimizing $\mathcal{L}_{refine}$.
\EndFor
\end{algorithmic}
\end{algorithm}

\subsection{Stage 1: Action-Conditioned Latent World Model}

Stage~1 learns the action-conditioned latent world model $W_\theta$. Given the current observation and an action prefix, it predicts the visual feature of the corresponding future observation. Simultaneously, and equally importantly, the same model produces the dynamics-aware current-state representation $c_t$ used by the Stage~2 policy. The key insight is that a model forced to anticipate how objects and the scene will evolve under actions must learn to attend to motion-relevant visual patterns---object boundaries, contact points, and task-relevant spatial relationships---and these patterns are naturally embedded in $c_t$. The design therefore separates two complementary outputs: an action-dependent future consequence $\hat I_f$ and an action-independent current representation $c_t$ shaped by the future-prediction objective.

\textbf{Input representation.} Let $E_v$ denote the frozen DINOv3 encoder \citep{simeoni2025dinov3}. The patch features from the current observation are concatenated as
\[
I_c=E_v(o_t)\in\mathbb{R}^{N\times d_v},
\]
where $N$ is the total number of visual tokens across multiple views. A visual projection $P_v$ maps $I_c$ to current-state tokens $C^0\in\mathbb{R}^{N\times d}$. The action projection $P_a$ maps an action chunk to tokens $A^0\in\mathbb{R}^{H\times d}$. A set of learnable future queries $Q\in\mathbb{R}^{N\times d}$, optimized as part of $W_\theta$, initializes $Q^0$ and provides the prediction slots from which the future feature is decoded. Learned positional embeddings are added independently to the three token groups.

\textbf{Variable-horizon action conditioning.} Instead of always predicting a fixed terminal observation, Stage~1 jointly varies the action prefix and its future target. For each sample, we draw
\[
n\sim\mathcal{U}\{1,\ldots,\min(H,L_t)\},
\]
where $L_t$ is the number of remaining environment steps. The truncated action chunk retains its first $n$ actions; the remaining positions are set to a stationary state, whose concrete form depends on the action space. For delta actions representing relative displacements, zeros indicate no motion. For absolute joint-space actions, the last valid pose $a_{t+n-1}$ is repeated across the remaining positions, so that the robot holds its configuration after the cutoff. Denoting the resulting chunk by $a_{t:t+H}^{(n)}$, its prediction target is synchronized to the same horizon:
\[
A^0=P_a\!\left(a_{t:t+H}^{(n)}\right),
\qquad
I_f^{(n)}=E_v(o_{t+n}).
\]
This paired truncation and target shift exposes the world model to consequences at different horizons and prevents it from associating every action input with a single fixed future frame. That is, this augmentation forces the world model to predict future visual states from partially observed action sequences and to infer how far into the future the truncated actions could carry the scene. As a result, the model cannot rely on the full action chunk as a shortcut and must develop a more robust understanding of action-conditioned dynamics, which in turn strengthens the quality of both the predicted future features $\hat I_f$ and the current latent $c_t$.

\textbf{Causal multimodal attention.} The model applies $L$ multimodal attention blocks to future-query tokens $Q^\ell$, current visual tokens $C^\ell$, and action tokens $A^\ell$. The three branches have independent QKV projections, output projections, and feed-forward networks, while attention is computed jointly over their concatenated keys and values. Information flow is controlled by the block visibility mask
\[
M=
\begin{bmatrix}
1 & 0 & 0\\
1 & 1 & 0\\
1 & 1 & 1
\end{bmatrix},
\]
whose rows and columns correspond to current-state, action tokens, and future-query tokens, respectively. Thus, future queries attend to all branches to combine the observed scene with the action prefix; current-state tokens attend only to themselves; and action tokens attend to the current-state and action branches. For block $\mathcal{B}_\theta^\ell$,
\[
(Q^{\ell+1},C^{\ell+1},A^{\ell+1})
=
\mathcal{B}_\theta^\ell(Q^\ell,C^\ell,A^\ell;M).
\]
The restricted current-state branch prevents action and future-query information from entering $C^\ell$, so its final output remains independent of the candidate action. Nevertheless, because the future-query branch uses $C^\ell$ as predictive context, gradients from future prediction train this branch to expose visual information useful for action-conditioned dynamics. We therefore define the final dynamics-shaped current representation as $c_t=C^L$.

\textbf{Latent future prediction.} A linear prediction head $P_f$ maps the final future-query tokens back to the frozen DINOv3 feature space:
\[
\hat I_f=P_f(Q^L),
\qquad
\mathcal{L}_{wm}
=
\left\|\hat I_f-I_f^{(n)}\right\|_1.
\]
The feature prediction $\hat I_f$ is later used as the imagined consequence of an action proposal, while $c_t$ provides the action-independent visual condition for Stage~2 action generation.

\subsection{Stage 2: World-Aware Action Generator}

\begin{figure}[t]
    \centering
    \includegraphics[width=0.9\linewidth]{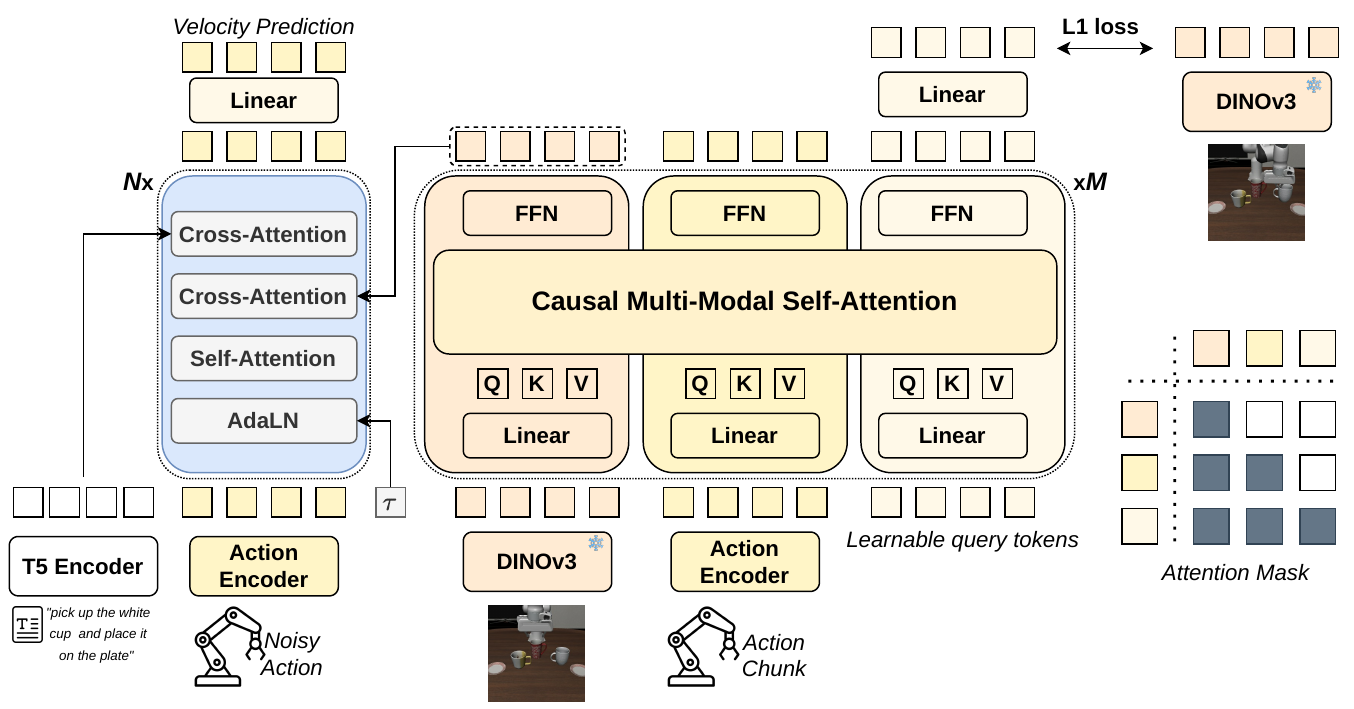}
    \caption{\textbf{Stage~2 world-aware action generation.} A time-conditioned Query Transformer takes the noised action chunk as query tokens. Each block performs action self-attention followed by cross-attention to the world-model representation $c_t$ and language tokens $T_e$, and predicts the flow velocity used to generate an initial action proposal.}
    \label{fig:cotraining}
\end{figure}

As illustrated in Figure~\ref{fig:cotraining}, Stage~2 couples the latent world model with a language-conditioned flow policy $\pi_\phi$. Its role is to transform the dynamics-shaped current representation $c_t$ learned in Stage~1 into an initial action proposal. Unlike Stage~1, Stage~2 uses the full expert action chunk, with episode-end padding where necessary, rather than a randomly truncated prefix. Passing this chunk through $W_\theta$ yields $c_t$, while a frozen T5 encoder \citep{raffel2020t5} produces instruction tokens that are mapped to the model dimension by a learned projection:
\[
\bar T_e=P_l(T_e).
\]
The world-model and language features remain separate conditioning sequences and are accessed through independent cross-attention modules.

\textbf{Flow-matching action queries.} We formulate action generation as conditional flow matching \citep{lipman2022flow}. To distinguish flow time from the environment index $t$, let $\tau\in[0,1]$ denote the flow timestep. During training, we sample
\[
\tau=\sigma(\xi),\qquad
\xi\sim\mathcal{N}(0,1),
\]
and Gaussian action noise $\epsilon\sim\mathcal{N}(0,I)$ with the same shape as the normalized expert chunk $a=a_{t:t+H}$. A point on the linear probability path between data and noise is
\[
x_\tau=(1-\tau)a+\tau\epsilon.
\]
Rather than using fixed learned queries, the policy directly embeds the noised action:
\[
X^0=P_a(x_\tau)+E_a,
\]
where $P_a$ is an action MLP and $E_a\in\mathbb{R}^{H\times d}$ is a learned action-position embedding. This preserves the temporal correspondence between each query token and its action-chunk position throughout denoising.

\textbf{Time-conditioned query transformer.} The flow timestep is encoded with sinusoidal features followed by an MLP, producing $e_\tau$. The policy then applies a sequence of Query Transformer blocks. In block $\mathcal{G}_\phi^\ell$, the action queries are updated in four steps:
\[
\begin{aligned}
\tilde X^\ell
&=\operatorname{SelfAttn}(X^\ell;e_\tau),\\
\bar X^\ell
&=\tilde X^\ell+
\operatorname{CrossAttn}_{v}(\tilde X^\ell,c_t),\\
\hat X^\ell
&=\bar X^\ell+
\operatorname{CrossAttn}_{l}(\bar X^\ell,\bar T_e),\\
X^{\ell+1}
&=\operatorname{FFN}(\hat X^\ell;e_\tau).
\end{aligned}
\]
The language cross-attention uses the text padding mask. Flow-time conditioning is injected into self-attention and the feed-forward network through adaptive layer normalization: $e_\tau$ predicts scale, shift, and residual-gating parameters for both transformations. The final action tokens are mapped to the velocity field by a linear head,
\[
v_\phi(x_\tau,c_t,T_e,\tau)=P_v(X^L).
\]

\textbf{Joint training objective.} Under the chosen interpolation, the target velocity is constant along the path:
\[
v^\star=\epsilon-a.
\]
The flow policy is trained with
\[
\mathcal{L}_{fm}
=
\left\|
v_\phi(x_\tau,c_t,T_e,\tau)-(\epsilon-a)
\right\|_2^2.
\]
We jointly optimize the proposal policy and latent world model using
\[
\mathcal{L}_{stage2}
=
\mathcal{L}_{fm}
+\lambda_{wm}\mathcal{L}_{wm},
\qquad \lambda_{wm}=0.1.
\]
The world-model term retains the future-prediction supervision while allowing $c_t$ to adapt to action generation. The implementation also supports freezing $W_\theta$ after Stage~1; in that setting, only $\pi_\phi$ is optimized and the world-model term is omitted.

\textbf{Initial proposal generation.} At inference, the policy first extracts $c_t$ from $W_\theta$ using a dummy zero-action input, as permitted by the action-independent current-state branch. Starting from Gaussian action noise $x_{\tau_S}\sim\mathcal{N}(0,I)$ at $\tau_S=1$, it follows a decreasing, sequence-length-shifted schedule to $\tau_0=0$, where $S$ is the number of integration steps. We use explicit Euler integration:
\[
x_{\tau_{s-1}}
=
x_{\tau_s}
+(\tau_{s-1}-\tau_s)
v_\phi(x_{\tau_s},c_t,T_e,\tau_s).
\]
The terminal sample defines the Stage~2 proposal,
\[
\hat a^0_{t:t+H}=x_{\tau_0},
\]
which is subsequently evaluated through the latent world model in Stage~3.

\subsection{Stage 3: Imagined-Interaction Action Refiner}

Stage~3 is the central step that turns the predictive and action-generation capabilities learned in Stages~1--2 into an interactive world-model policy. Before this stage, the world model supplies future supervision and a dynamics-shaped representation for action generation. Stage~3 introduces the missing feedback path: the policy's own action proposal is returned to the world model, and the consequence imagined for that proposal is used to produce the final action. The world model therefore participates directly in a single control decision rather than remaining only a training objective or visual backbone.

\textbf{Proposal-conditioned imagination.} We freeze the world model $W_\theta$ and proposal policy $\pi_\phi$, and train only the refiner $R_\psi$. Given $I_c$ and $T_e$, the frozen Stage~2 pipeline first extracts the current representation with a dummy zero-action input and generates an initial proposal through flow integration:
\[
c_t=W_\theta(I_c,a_{\mathrm{dummy}}),
\qquad
\hat a^0_{t:t+H}=\pi_\phi(c_t,T_e).
\]
The proposal is then fed back into the frozen world model to obtain its action-conditioned latent consequence:
\[
(\hat I_f,c_t)=W_\theta(I_c,\hat a^0_{t:t+H}).
\]
Here, $\hat I_f$ is the same latent future representation learned through DINO-feature prediction in Stage~1, now conditioned on the policy's own proposal rather than an expert action. Because the current-state branch of $W_\theta$ cannot attend to action or future-query tokens, the returned $c_t$ remains a function of the current observation alone; only $\hat I_f$ is conditioned on the proposal. All condition extraction is performed without gradient propagation into $W_\theta$ or $\pi_\phi$.

\textbf{Tri-branch interaction refiner.} The refiner receives three aligned token groups:
\[
C^0=c_t+E_c,\qquad
F^0=\hat I_f+E_f,\qquad
U^0=P_r(\hat a^0_{t:t+H})+E_a,
\]
where $P_r$ projects the proposal into the model dimension and $E_c,E_f,E_a$ are learned positional embeddings. These branches expose the three quantities required for consequence-aware correction: $C^0$ represents the currently observed state, $F^0$ represents what the proposed action is predicted to cause, and $U^0$ represents the action currently under consideration.

The refiner applies $L_r$ tri-branch multimodal attention blocks. Each branch has independent QKV projections, output projection, and feed-forward network, while attention is computed jointly over the concatenated current, future, and proposal tokens. Unlike the structured visibility mask in Stage~1, Stage~3 uses no attention mask, allowing all three branches to interact:
\[
(F^{\ell+1},C^{\ell+1},U^{\ell+1})
=
\mathcal{R}_\psi^\ell(F^\ell,C^\ell,U^\ell).
\]
This full interaction lets each proposal token compare its intended motion against both the present scene and its predicted outcome. Stage~3 introduces neither a flow timestep nor adaptive layer normalization, and it does not run a second denoising process. A linear action head directly maps the final proposal branch to the refined action:
\[
\hat a_{t:t+H}=P_{\mathrm{out}}(U^{L_r}).
\]
Accordingly, the refiner does not predict an explicit verification score or a residual offset. In Envision--Verify--Act, \emph{verify} denotes using the imagined consequence to assess and directly rewrite the proposed action.

\textbf{Refinement supervision.} Stage~3 does not receive the ground-truth future observation as an input. Its future feedback is generated entirely by the frozen world model from the policy's own proposal, matching the information available at deployment. Only the refined action is supervised. With a validity mask $m_h\in\{0,1\}$ for each action step, we minimize the mean absolute error over valid action elements:
\[
\mathcal{L}_{refine}
=
\frac{
\sum_{h=1}^{H}m_h
\left\|\hat a_{t+h}-a_{t+h}\right\|_1
}{
d_a\sum_{h=1}^{H}m_h
}.
\]
This objective trains $R_\psi$ to determine how the proposal should change when its predicted consequence is considered. Stage~3 thereby closes the proposal--imagination--refinement loop: the world model becomes a proposal-conditioned action-feedback module inside the policy, which is the defining mechanism of $\omega$-EVA's latent interactive world-model paradigm.

\subsection{Inference: Envision--Verify--Act}

At deployment, $\omega$-EVA performs one Envision--Verify--Act interaction before committing an action chunk. Given the current observation $o_t$ and instruction $l$, the frozen visual and language encoders first produce
\[
I_c=E_v(o_t),
\qquad
T_e=E_l(l).
\]
The world model is then evaluated with a dummy zero-action input to extract the action-independent current representation,
\[
c_t=W_\theta(I_c,a_{\mathrm{dummy}}).
\]
Conditioned on $(c_t,T_e)$, the proposal policy initializes $x_{\tau_S}\sim\mathcal{N}(0,I)$ at $\tau_S=1$ and integrates its learned velocity field along the decreasing shifted schedule $\tau_S>\cdots>\tau_0=0$:
\[
x_{\tau_{s-1}}
=
x_{\tau_s}
+(\tau_{s-1}-\tau_s)
v_\phi(x_{\tau_s},c_t,T_e,\tau_s).
\]
The terminal sample $\hat a^0_{t:t+H}=x_{\tau_0}$ is the action considered by the world model.

\textbf{Envision and verify.} $\omega$-EVA feeds this specific proposal back into the frozen world model to predict its latent consequence, and the refiner uses that consequence as action feedback:
\[
\hat I_f
=
W_\theta(I_c,\hat a^0_{t:t+H}),
\qquad
\hat a_{t:t+H}
=
R_\psi(c_t,\hat I_f,\hat a^0_{t:t+H}).
\]
Here, \emph{verify} is a single consequence-aware refinement rather than reward evaluation, an explicit verification score, or multi-step planning. The system neither observes the true future nor decodes $\hat I_f$ into a future video; all consequence reasoning remains in the latent feature space.

\textbf{Receding-horizon execution.} After refinement, the robot executes the first $K\leq H$ actions from $\hat a_{t:t+H}$, receives a new observation $o_{t+K}$, and repeats the full interaction. Setting $K=1$ gives step-wise closed-loop control, while $K>1$ gives action-chunk execution between replanning steps. The execution prefix is selected by the deployment protocol rather than fixed by the model. Algorithm~\ref{alg:omega_infer} summarizes this deployment loop.

\begin{algorithm}[t]
\caption{Envision--Verify--Act inference}
\label{alg:omega_infer}
\begin{algorithmic}[1]
\Require Current observation $o_t$, instruction $l$, execution prefix $K\leq H$, encoders $E_v,E_l$, $W_\theta$, $\pi_\phi$, and $R_\psi$
\Ensure Refined action chunk $\hat a_{t:t+H}$
\State Encode $I_c=E_v(o_t)$ and $T_e=E_l(l)$.
\State Extract $c_t$ from $W_\theta(I_c,a_{\mathrm{dummy}};Q)$ with a dummy zero action.
\State Sample $x_{\tau_S}\sim\mathcal{N}(0,I)$ with $\tau_S=1$.
\For{$s=S,S-1,\ldots,1$}
    \State $v_s\gets v_\phi(x_{\tau_s},c_t,T_e,\tau_s)$.
    \State $x_{\tau_{s-1}}\gets x_{\tau_s}+(\tau_{s-1}-\tau_s)v_s$.
\EndFor
\State Set proposal $\hat a^0_{t:t+H}\gets x_{\tau_0}$.
\State \textbf{Envision:} $\hat I_f\gets W_\theta(I_c,\hat a^0_{t:t+H};Q)$.
\State \textbf{Verify:} $\hat a_{t:t+H}\gets R_\psi(c_t,\hat I_f,\hat a^0_{t:t+H})$.
\State Execute the first $K$ actions of $\hat a_{t:t+H}$, observe $o_{t+K}$, and repeat.
\end{algorithmic}
\end{algorithm}

Thus, $\omega$-EVA retains explicit test-time consequence reasoning without invoking a separate pixel-level simulator: the world model acts as an internal, proposal-conditioned feedback module within each control decision.

%% file: chapters/4_Experiments.tex
\section{Experiments}

We evaluate $\omega$-EVA on three simulated manipulation benchmarks: LIBERO~\citep{liu2023libero}, LIBERO-PLUS~\citep{fei2025liberoplus}, and RoboTwin~2.0~\citep{chen2025robotwin2}. Together, they cover single-arm and bimanual control, long-horizon task execution, and robustness to visual, linguistic, and environmental perturbations. We report task success rate, robot-pretraining usage, and core model size. Our evaluation focuses on whether the compact $\omega$-EVA architecture remains competitive without robot pretraining and whether Stage~3 consistently improves its Stage~2 proposal policy through proposal-conditioned imagined consequences.

\subsection{Experimental Setup}

\paragraph{Implementation.}
For LIBERO and LIBERO-PLUS, $\omega$-EVA uses an agent-view image and a wrist-view image, each resized to $256\times256$ and encoded independently before their visual tokens are concatenated. The action horizon is $H=16$ and the action dimension is $7$. RoboTwin~2.0 provides left-wrist, right-wrist, and head-camera observations; following the benchmark setup used by Fast-WAM \citep{yuan2026fastwam}, we stitch the three views into one image before visual encoding. The action horizon remains $H=16$, with a $14$-dimensional action for bimanual control.

For $\omega$-EVA, the Stage~1 latent world model contains 12 decoupled multimodal-attention blocks with hidden dimension 1024 and 8 attention heads. The Stage~2 flow policy uses 12 Query Transformer blocks and 5 Euler integration steps at inference. The Stage~3 refiner contains 12 tri-branch joint-attention blocks and directly predicts the refined action chunk.

$\omega$-EVA uses \emph{no robot pretraining}. The world model, action policy, and refiner are initialized and trained only on the training data of the corresponding benchmark; they do not load robot-policy, VLA, robot-trajectory, or robot-video pretraining. We use frozen DINOv3 and text encoders solely to extract generic visual and language features, following the architecture in Section~3. Accordingly, ``Robot Pretrain'' in the following tables denotes pretraining on additional robot interaction data, rather than the use of frozen generic representation encoders.

The Stage~2 world-model--policy stack contains approximately 0.8B parameters, and adding the Stage~3 refiner increases $\omega$-EVA to approximately 1.2B parameters. Both counts include the frozen DINOv3 visual encoder but exclude the frozen T5 text encoder. For baselines, we report the main policy or world-action-model size stated by the corresponding work. Since different works may account for frozen or auxiliary encoders differently, the parameter column indicates model scale rather than a strictly standardized parameter-count benchmark.

\paragraph{Training details.}
All three stages are trained with AdamW, an initial learning rate of $10^{-4}$, a cosine-annealed final learning rate of $10^{-6}$, weight decay $10^{-6}$, and $(\beta_1,\beta_2)=(0.9,0.95)$. We use 500 warmup iterations, bfloat16 mixed precision, and Distributed Data Parallel training. Our main runs use 16 NVIDIA H100 GPUs with 80\,GB memory each and a total batch size of 1024. Stage~1, Stage~2, and Stage~3 are trained for 50, 30, and 20 epochs, respectively. We evaluate checkpoints from multiple epochs through repeated simulator rollouts and report the best rollout result for each stage.

\subsection{Simulation Benchmarks}

\paragraph{LIBERO.}
LIBERO \citep{liu2023libero} comprises four standard suites---Spatial, Object, Goal, and Long---that test spatial reasoning, object interaction, goal interpretation, and long-horizon execution. Each task is initialized from benchmark-provided states and evaluated with the official simulator horizon. Table~\ref{tab:libero_results} compares $\omega$-EVA with published VLA and world-action-model results. The reported baselines differ substantially in their pretraining data and model scale, so we expose their robot-pretraining status rather than treating the table as a strictly controlled ranking.

\begin{table}[t]
\centering
\caption{\textbf{Success rates (\%) on LIBERO.} Baseline values are reported by Fast-WAM \citep{yuan2026fastwam} and the cited original works. ``Robot Pretrain'' indicates additional robot interaction data before benchmark training. Params excludes T5 for $\omega$-EVA and follows the published main-model size for baselines. Bold marks the best average within each robot-pretraining group.}
\label{tab:libero_results}
\resizebox{\linewidth}{!}{%
\begin{tabular}{lccccccc}
\toprule
Method & Core Params & Robot Pretrain & Spatial & Object & Goal & Long & Avg. \\
\midrule
OpenVLA~\citep{kimOpenVLAOpenSourceVisionLanguageAction2024} & 7B   & Yes & 84.7 & 88.4 & 79.2 & 53.7 & 76.5 \\
$\pi_0$~\citep{black2024pi_0}                             & 3.3B & Yes & 96.8 & 98.8 & 95.8 & 85.2 & 94.1 \\
$\pi_{0.5}$~\citep{intelligence2025pi_}                   & 3.3B & Yes & 98.8 & 98.2 & 98.0 & 92.4 & 96.9 \\
LingBot-VA~\citep{li2026lingbotva}                         & 5.3B & Yes & 98.5 & 99.6 & 97.2 & 98.5 & \textbf{98.5} \\
Cosmos Policy~\citep{kim2026cosmos}                    & 2B   & Yes & 98.1 & 100.0 & 98.2 & 97.6 & \textbf{98.5} \\
Motus~\cite{bi2026motus}                                  & 8B   & Yes & 96.8 & 99.8 & 96.6 & 97.6 & 97.7 \\
VLA-JEPA~\cite{sun2026vlajepa}                                  & 3B   & Yes & 96.2 & 99.6 & 97.2 & 95.8 & 97.2 \\
\midrule
VLA-JEPA w/o human videos~\cite{sun2026vlajepa}                                 & 3B   & No & 94.8 & 99.6 & 95.8 & 94.0 & 96.1 \\
Fast-WAM~\citep{yuan2026fastwam}                           & 6B   & No  & 98.2 & 100.0 & 97.0 & 95.2 & 97.6 \\
\textbf{$\omega$-EVA Stage~2 (Ours)} & 0.8B & No & 98.8 & 99.4 & 97.6 & 95.8 & 97.9 \\
\textbf{$\omega$-EVA Stage~3 (Ours)} & 1.2B & No & 99.0 & 99.8 & 98.2 & 97.4 & \textbf{98.6} \\
\bottomrule
\end{tabular}%
}
\end{table}

Without robot pretraining, the compact 0.8B Stage~2 model reaches an average success rate of $97.9\%$. The complete Envision--Verify--Act pipeline raises this result by $0.7$ points to $98.6\%$, improving all four suites and yielding its largest gain of $1.6$ points on LIBERO-Long. The 1.2B Stage~3 model achieves the highest average in the table despite using only benchmark training data, outperforming substantially larger robot-pretrained and benchmark-only models. These results establish both a favorable performance--scale trade-off and a consistent benefit from adding the full Stage~3 pipeline to the same proposal policy.

\paragraph{LIBERO-PLUS robustness and transfer.}
LIBERO-PLUS \citep{fei2025liberoplus} extends LIBERO with seven controlled perturbation categories spanning camera, robot appearance, language, lighting, background, sensor noise, and scene layout. We evaluate $\omega$-EVA in two settings. For zero-shot transfer, the model is trained only on LIBERO and directly evaluated on LIBERO-PLUS. For benchmark-specific training, all trainable $\omega$-EVA modules are trained on the LIBERO-PLUS split, still without any additional robot pretraining. Table~\ref{tab:liberoplus_results} includes representative baselines from the LIBERO-PLUS paper and reports the training data used by each $\omega$-EVA variant.

\begin{table}[t]
\centering
\caption{\textbf{Success rates (\%) under LIBERO-PLUS perturbations.} Published baseline values are from \citet{fei2025liberoplus}, \citet{sun2026vlajepa}, and the cited original works. Params excludes T5 for $\omega$-EVA and follows published main-model sizes for baselines. Bold marks the best average within each training-data and robot-pretraining group.}
\label{tab:liberoplus_results}
\resizebox{\linewidth}{!}{%
\begin{tabular}{lclccccccccc}
\toprule
Method & Core Params & Training Data & Robot Pretrain & Camera & Robot & Language & Light & Background & Noise & Layout & Avg. \\
\midrule
OpenVLA~\citep{kimOpenVLAOpenSourceVisionLanguageAction2024} & 7B   & LIBERO & Yes & 0.8  & 3.5  & 23.0 & 8.1  & 34.8 & 15.2 & 28.5 & 15.6 \\
$\pi_0$~\citep{black2024pi_0}                             & 3.3B & LIBERO & Yes & 13.8 & 6.0  & 58.8 & 85.0 & 81.4 & 79.0 & 68.9 & 53.6 \\
$\pi_0$-FAST~\citep{pertsch2025pifast}                    & 3.3B & LIBERO & Yes & 65.1 & 21.6 & 61.0 & 73.2 & 73.2 & 74.4 & 68.8 & 61.6 \\
OpenVLA-OFT~\cite{kim2025openvla_oft}                & 7B   & LIBERO      & Yes & 56.4  & 31.9  & 79.5  & 88.7  & 93.3  & 75.8  & 74.2  & 69.6 \\
VLA-JEPA~\cite{sun2026vlajepa}                & 3B   & LIBERO      & Yes & 63.3 & 67.1  & 85.4  & 95.6  & 93.6  & 66.3  & 85.1  & \textbf{79.5} \\
\midrule
VLA-JEPA w/o human videos~\cite{sun2026vlajepa}                & 3B   & LIBERO      & No & 40.3  & 55.7  & 72.9  & 88.2  & 70.5  & 38.2  & 74.6  & 62.9 \\
Fast-WAM~\citep{yuan2026fastwam}               & 6B   & LIBERO      & No & 16.4  & 44.5  & 68.9  & 78.2  & 53.7  & 37.7  & 60.7  & 51.5 \\
\textbf{$\omega$-EVA Stage~2 (Ours)}                     & 0.8B & LIBERO      & No  & 59.5 & 58.8 & 66.4 & 97.1 & 74.5 & 74.1 & 76.7 & 71.3 \\
\textbf{$\omega$-EVA Stage~3 (Ours)}                     & 1.2B & LIBERO      & No  & 62.8 & 61.6 & 64.4 & 96.7 & 76.9 & 74.1 & 76.8 & \textbf{72.2} \\
\midrule
LIBERO-PLUS post-trained~\citep{fei2025liberoplus}   & 7B   & LIBERO-PLUS & Yes & 92.8  & 30.3  & 85.8  & 94.9  & 93.9  & 89.3  & 77.6  & 79.5 \\
\textbf{$\omega$-EVA Stage~2 (Ours)}                     & 0.8B & LIBERO-PLUS & No  & 92.0 & 62.8 & 66.1 & 93.5 & 92.2 & 90.8 & 76.7 & 81.2 \\
\textbf{$\omega$-EVA Stage~3 (Ours)}                     & 1.2B & LIBERO-PLUS & No  & 95.3 & 65.2 & 66.1 & 94.5 & 93.9 & 94.4 & 79.5 & \textbf{83.4} \\
\bottomrule
\end{tabular}%
}
\end{table}

Under zero-shot transfer from LIBERO, the complete Stage~3 pipeline improves the average by $0.9$ points, from $71.3\%$ to $72.2\%$. Among methods without robot pretraining, $\omega$-EVA exceeds VLA-JEPA and Fast-WAM by $9.3$ and $20.7$ points, respectively, while using a smaller model. The Stage~3 gain is not uniform across categories: Camera, Robot, Background, and Layout improve, Noise is unchanged after rounding, while Language and Light decrease slightly. With LIBERO-PLUS benchmark training, Stage~3 produces a larger $2.2$-point gain, from $81.2\%$ to $83.4\%$, improving six of seven perturbation categories and matching Language. These results support a measured robustness claim: the full Envision--Verify--Act pipeline consistently improves aggregate performance and benefits most visual perturbations, without implying universal gains under every shift.

\paragraph{RoboTwin 2.0.}
RoboTwin~2.0 \citep{chen2025robotwin2} evaluates bimanual manipulation across clean and domain-randomized conditions. It complements LIBERO by increasing action dimensionality, camera coverage, and sensitivity to visual and physical variation. Table~\ref{tab:robotwin_results} compares our result with published VLA and world-action-model results under the benchmark's clean and randomized task groups.

\begin{table}[t]
\centering
\caption{\textbf{Success rates (\%) on RoboTwin~2.0.} Baseline values are reported by Fast-WAM \citep{yuan2026fastwam} and the cited original works. Params excludes T5 for $\omega$-EVA and follows published main-model sizes for baselines. Bold marks the best average within each robot-pretraining group. $\omega$-EVA uses only RoboTwin~2.0 training data.}
\label{tab:robotwin_results}
\resizebox{\linewidth}{!}{%
\begin{tabular}{lccccc}
\toprule
Method & Core Params & Robot Pretrain & Clean & Randomized & Avg. \\
\midrule
$\pi_0$~\citep{black2024pi_0}             & 3.3B & Yes & 65.9 & 58.4 & 62.2 \\
$\pi_{0.5}$~\citep{intelligence2025pi_}   & 3.3B & Yes & 82.7 & 76.8 & 79.8 \\
LingBot-VA~\citep{li2026lingbotva}         & 5.3B & Yes & 92.9 & 91.5 & 92.2 \\
Motus~\citep{bi2026motus}                  & 8B   & Yes & 88.7 & 87.0 & 87.8 \\
GigaWorld-Policy~\citep{ye2026gigaworldpolicy} & 5B   & Yes & 87.0 & 85.0 & 86.0 \\
Being-H0.7~\citep{luo2026being07}          & 3B   & Yes & 90.2 & 89.6 & 89.9 \\
MotuBrain~\citep{team2026motubrain}        & 8B   & Yes & 95.8 & 96.1 & \textbf{96.0} \\
\midrule
Motus w/o Pretrain~\citep{bi2026motus}     & 8B   & No & 77.6 & 77.0 & 77.3 \\
MotuBrain w/o Pretrain~\citep{team2026motubrain}        & 8B   & No & 91.5 & 91.3 & 91.4 \\
Fast-WAM~\citep{yuan2026fastwam}           & 6B   & No  & 91.9 & 91.8 & 91.8 \\
\textbf{$\omega$-EVA Stage~2 (Ours)} & 0.8B & No & 91.4 & 89.4 & 90.4 \\
\textbf{$\omega$-EVA Stage~3 (Ours)} & 1.2B & No & 93.3 & 91.1 & \textbf{92.2} \\
\bottomrule
\end{tabular}
}
\end{table}

The complete Stage~3 pipeline improves the RoboTwin average by $1.4$ points, from $88.9\%$ to $90.3\%$, with gains in both clean and randomized settings. $\omega$-EVA does not achieve the highest absolute RoboTwin score; instead, it offers a favorable performance--scale--data trade-off. Its 1.2B model surpasses the reported Motus and Being-H0.7 results and remains within $1.5$ points of the 6B Fast-WAM model, while using no robot pretraining. We therefore characterize $\omega$-EVA as compact and competitive rather than state of the art. The consistent Stage~3 gain further shows that the full interaction pipeline transfers from single-arm LIBERO tasks to higher-dimensional bimanual control.

\subsection{Ablation Studies}

Unless otherwise specified, all ablations are conducted on LIBERO using the same visual inputs, action horizon, training configuration, and rollout protocol as the main experiment. Each study changes only the component under examination. We first analyze whether future-prediction training produces a dynamics-aware yet action-independent current representation, then evaluate latent future fidelity and isolate the source of the Stage~3 improvement.

\subsubsection{Dynamics-Aware Current Representation and Action Invariance}

Stage~1 is designed to produce two complementary outputs: an action-conditioned future prediction $\hat I_f$ and an action-independent current representation $c_t$ shaped by future-prediction supervision. We examine both the spatial structure learned by the current branch and whether the decoupled attention mask prevents ground-truth actions from leaking into the Stage~2 policy condition.

\begin{figure}[H]
\centering
\begin{minipage}[t]{0.235\linewidth}
    \centering
    \includegraphics[width=\linewidth]{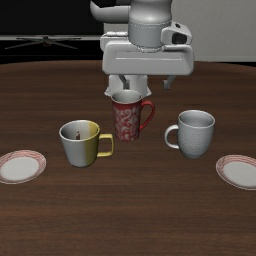}\\[-1mm]
    {\scriptsize (a) Current observation}
\end{minipage}\hfill
\begin{minipage}[t]{0.235\linewidth}
    \centering
    \includegraphics[width=\linewidth]{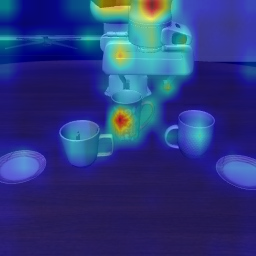}\\[-1mm]
    {\scriptsize (b) DINOv3}
\end{minipage}\hfill
\begin{minipage}[t]{0.235\linewidth}
    \centering
    \includegraphics[width=\linewidth]{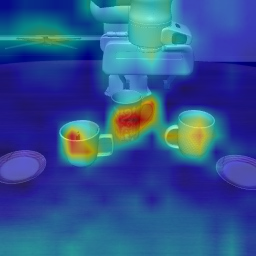}\\[-1mm]
    {\scriptsize (c) Stage~1 $c_t$}
\end{minipage}\hfill
\begin{minipage}[t]{0.235\linewidth}
    \centering
    \includegraphics[width=\linewidth]{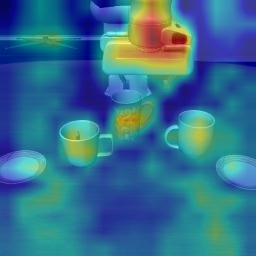}\\[-1mm]
    {\scriptsize (d) Stage~2 $c_t$}
\end{minipage}
\caption{\textbf{Spatial activation of the current visual representation.} Compared with the generic DINOv3 features in (b), future-prediction training in (c) redistributes activation toward the robot end effector, manipulated object, and nearby interaction regions. The Stage~2 co-trained world model in (d) retains these dynamics-relevant cues while adapting the representation for action generation. The heatmaps are qualitative representation diagnostics and do not by themselves establish globally superior features.}
\label{fig:current_representation}
\end{figure}

\paragraph{Spatial representation analysis.}
We visualize the spatial activation of the frozen DINOv3 features, the Stage~1 current representation, and the current representation after Stage~2 co-training. For a patch token $z_p$, its activation is defined as
\[
s_p=\left\|z_p\right\|_2.
\]
The patch scores are reshaped to their spatial grid, upsampled to the input resolution, and overlaid on the same current observation. We use the same colormap and normalize each heatmap independently; therefore, Figure~\ref{fig:current_representation} compares where each representation concentrates its activation, rather than absolute activation magnitudes across models.

The frozen DINOv3 map emphasizes generic visually salient regions. After Stage~1, activation becomes more concentrated around the end effector, the manipulated cup, and nearby objects whose spatial relations can change under robot actions. Stage~2 co-training preserves these interaction-relevant regions while adapting $c_t$ to condition the flow policy. This progression is consistent with the intended role of future prediction: it shapes the current branch to expose visual structure useful for scene dynamics without requiring the branch itself to observe an action.

\paragraph{Policy performance with current-representation variants.}
We complement the qualitative analysis with two controlled Stage~2 ablations. All variants use the same policy architecture, training configuration, and rollout protocol, and differ only in the visual representation supplied to the policy and whether the Stage~1 world model is updated during Stage~2. \textbf{Raw frozen DINOv3 features} bypasses Stage~1 and provides the generic visual features to the Stage~2 policy through a dimension-matched interface. \textbf{Stage~1 $c_t$ (frozen)} uses the current representation learned through future-prediction training, while freezing the world model throughout Stage~2. The full variant uses the same Stage~1 initialization but co-trains the world model and policy with the Stage~2 objective.

\begin{table}[H]
\centering
\caption{\textbf{Stage~2 policy performance with current-representation variants on LIBERO.} The raw-feature variant bypasses Stage~1, the frozen variant retains the dynamics-shaped Stage~1 representation without policy-driven adaptation, and the full variant co-trains the world model and policy. Success rates are reported in percent.}
\label{tab:current_representation_ablation}
% \resizebox{\linewidth}{!}{%
\begin{tabular}{lccccc}
\toprule
Current representation & Spatial & Object & Goal & Long & Avg. \\
\midrule
Raw frozen DINOv3 features & 96.4 & 97.0 & 93.2 & 81.8 & 92.1 \\
Stage~1 $c_t$ (frozen)     & 98.2 & 99.8 & 94.6 & 91.0 & 95.9 \\
Stage~2 co-trained $c_t$   & 98.8 & 99.4 & 97.6 & 95.8 & \textbf{97.9} \\
\bottomrule
\end{tabular}%
% }
\end{table}

Replacing raw DINOv3 features with the frozen Stage~1 representation raises the average success rate from $92.1\%$ to $95.9\%$, a $3.8$-point gain, with particularly large improvements on LIBERO-Long. This result provides behavioral evidence that future-prediction training shapes $c_t$ into a more useful policy condition than the generic visual features alone. Allowing the representation to adapt jointly with the policy further increases the average to $97.9\%$, an additional $2.0$-point gain. Co-training does not improve every suite individually---Object decreases slightly from $99.8\%$ to $99.4\%$---but it yields a clear aggregate improvement and substantial gains on Goal and Long. Together with the spatial analysis, these results support the claim that future-prediction-shaped current features and their policy-driven co-adaptation improve control performance; they also establish that $c_t$ is a representation of the underlying physical dynamics.

% To extend this qualitative example to the full evaluation set, we define an interaction region of interest using the union of LIBERO simulator segmentation masks for the robot and task-relevant objects. Patch activation is normalized by the total activation in each image. Table~\ref{tab:representation_localization} will report the fraction of activation inside this region, the precision of the top $10\%$ most active patches, and the remaining background mass.

% \begin{table}[H]
% \centering
% \caption{\textbf{Localization of current-representation activation on LIBERO.} The interaction region is the union of robot and task-relevant object masks. Results for the full evaluation set are pending.}
% \label{tab:representation_localization}
% \resizebox{\linewidth}{!}{%
% \begin{tabular}{lccc}
% \toprule
% Representation & Interaction-region mass $\uparrow$ & Top-10\% patch precision $\uparrow$ & Background mass $\downarrow$ \\
% \midrule
% DINOv3                    & TBD & TBD & TBD \\
% Stage~1 $c_t$             & TBD & TBD & TBD \\
% Stage~2 co-trained $c_t$  & TBD & TBD & TBD \\
% \bottomrule
% \end{tabular}%
% }
% \end{table}

\paragraph{Action-invariance sanity check.}
During Stage~2 co-training, the world-model forward pass receives the expert action to retain future-prediction supervision. This could create target leakage if action information entered the current branch used by the policy. We test this directly by holding the current images fixed and extracting
\[
c_t^{(a)}=W_\theta^{c}(I_c,a),
\qquad
\Delta(a,a')=c_t^{(a)}-c_t^{(a')},
\]
where $W_\theta^{c}$ denotes the current-state output of the world model. We compare the expert chunk with a dummy zero chunk, Gaussian random actions, and expert chunks shuffled across the batch.

\begin{table}[H]
\centering
\caption{\textbf{Action-invariance of $c_t$ on the LIBERO evaluation set.} Differences are computed element-wise between current representations extracted from identical images under different action inputs. Expert, dummy, random, and batch-shuffled actions all produce numerically identical $c_t$ values.}
\label{tab:action_invariance}
\begin{tabular}{lccc}
\toprule
Action comparison & Mean abs. diff. $\downarrow$ & Max abs. diff. $\downarrow$ & Cosine similarity $\uparrow$ \\
\midrule
Expert vs.\ dummy    & 0.000 & 0.000 & 1.000 \\
Expert vs.\ random   & 0.000 & 0.000 & 1.000 \\
Expert vs.\ shuffled & 0.000 & 0.000 & 1.000 \\
\bottomrule
\end{tabular}
\end{table}

Across the full LIBERO evaluation set, replacing the expert chunk with dummy, random, or batch-shuffled actions leaves $c_t$ numerically unchanged: all comparisons have zero mean and maximum absolute difference and unit cosine similarity. This confirms that the action supplied for Stage~2 future supervision does not enter the policy condition. The result follows the decoupled visibility mask, under which current-state tokens attend only to the current branch, and rules out equality being a special case of the dummy input. Importantly, this invariance applies only to $c_t$: the future-query branch $\hat I_f$ remains action-conditioned by design.

\subsubsection{Action-Conditioned Latent Future Fidelity}

\paragraph{Evaluation protocol.}
Our world model predicts future DINOv3 features rather than pixels. To make these latent consequences observable, we train a single diagnostic decoder $D$ to map frozen DINOv3 features back to image space. The decoder is used only for this analysis and is not part of policy training or inference. For each valid LIBERO sequence, we pair the current observation $o_t$ with the future observation $o_{t+16}$ after one complete action chunk, discarding episode tails shorter than 16 steps.

We evaluate four conditions. First, $D(E_v(o_{t+16}))$ reconstructs the true future feature and measures the information preserved by the DINOv3--decoder projection. Second, the world model predicts a future latent from the current observation and the ground-truth action chunk. Third and fourth, we replace the ground-truth action with the Stage~2 proposal and Stage~3 refined action, respectively. All predicted latents are visualized through the same decoder and compared directly with the original future observation $o_{t+16}$. We report the Structural Similarity Index Measure (SSIM), a paired image metric that evaluates structural agreement and is higher when the decoded prediction more closely matches its target, and the Fr\'echet Inception Distance (FID), a set-level metric that compares the feature distributions of decoded predictions and real future images and is lower when the distributions are closer. Because the decoder and metrics use the same LIBERO data, this experiment is a representation diagnostic rather than a held-out reconstruction or generalization evaluation.

\begin{table}[H]
\centering
\caption{\textbf{Action-conditioned future-latent fidelity on LIBERO.} All decoded predictions are evaluated against the original future observation $o_{t+16}$. SSIM measures paired structural similarity, whereas FID compares the distributions of the corresponding decoded and real image sets. Bold marks the better result between Stage~2 and Stage~3.}
\label{tab:latent_fidelity}
% \resizebox{\linewidth}{!}{%
\begin{tabular}{lcc}
\toprule
Condition & SSIM vs. raw GT $\uparrow$ & FID vs. raw GT $\downarrow$ \\
\midrule
DINO reconstruction ceiling & 0.9700 & 7.10 \\
GT-action imagination       & 0.9569 & 7.40 \\
Stage~2 proposal imagination & 0.9520 & 7.41 \\
Stage~3 refined-action imagination & \textbf{0.9562} & \textbf{7.40} \\
\bottomrule
\end{tabular}%
% }
\end{table}

\begin{figure}[t]
\centering
\begin{minipage}[t]{0.31\linewidth}
    \centering
    \includegraphics[width=\linewidth]{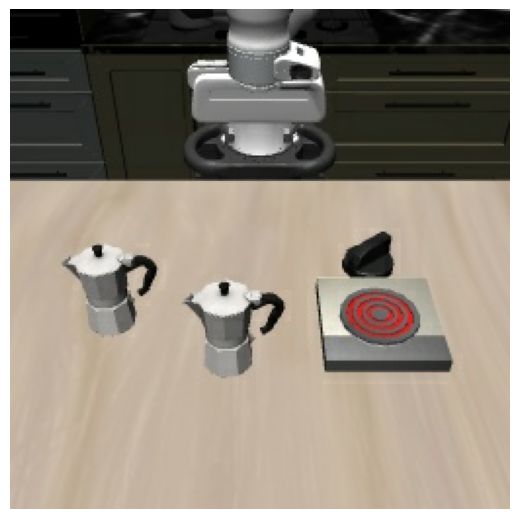}\\[-1mm]
    {\scriptsize (a) Current observation}
\end{minipage}\hfill
\begin{minipage}[t]{0.31\linewidth}
    \centering
    \includegraphics[width=\linewidth]{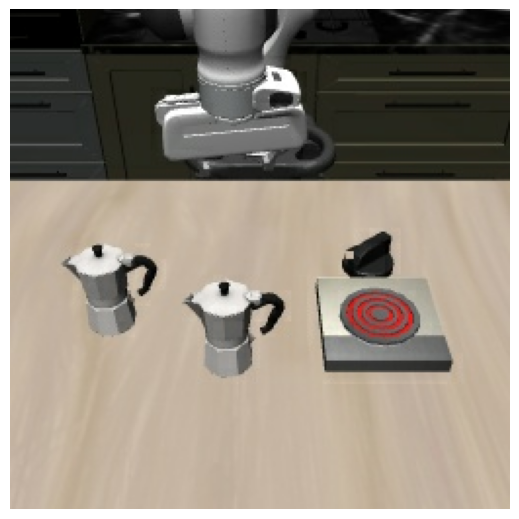}\\[-1mm]
    {\scriptsize (b) Raw GT future}
\end{minipage}\hfill
\begin{minipage}[t]{0.31\linewidth}
    \centering
    \includegraphics[width=\linewidth]{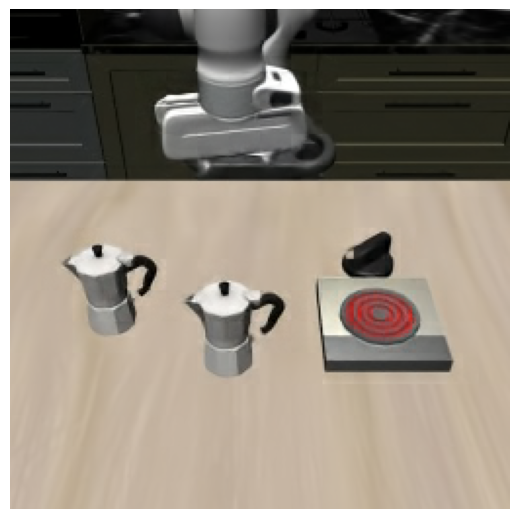}\\[-1mm]
    {\scriptsize (c) Decoded GT future}
\end{minipage}

\vspace{2mm}

\begin{minipage}[t]{0.31\linewidth}
    \centering
    \includegraphics[width=\linewidth]{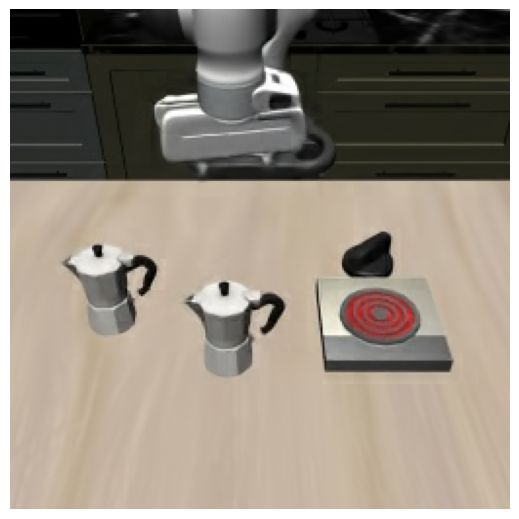}\\[-1mm]
    {\scriptsize (d) GT-action imagination}
\end{minipage}\hfill
\begin{minipage}[t]{0.31\linewidth}
    \centering
    \includegraphics[width=\linewidth]{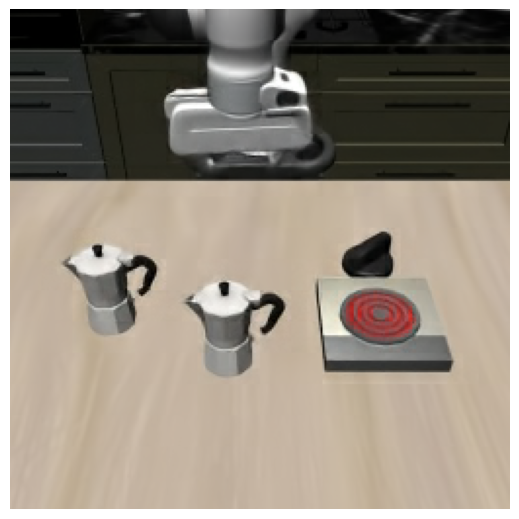}\\[-1mm]
    {\scriptsize (e) Stage~2 proposal}
\end{minipage}\hfill
\begin{minipage}[t]{0.31\linewidth}
    \centering
    \includegraphics[width=\linewidth]{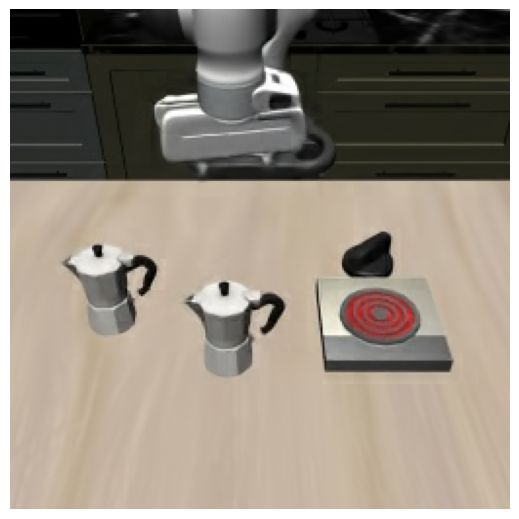}\\[-1mm]
    {\scriptsize (f) Stage~3 refinement}
\end{minipage}
\caption{\textbf{Decoded action-conditioned future latents.} All latent predictions are rendered by the same diagnostic decoder. The GT-action prediction in (d) closely follows the decoded future target in (c), indicating that the world model captures action-conditioned scene evolution. Compared with the Stage~2 proposal in (e), the Stage~3 refined action in (f) produces a future whose robot pose and scene structure more closely match the target and GT-action reference. Decoder smoothing reflects the diagnostic projection and is not pixel generation performed by the policy.}
\label{fig:latent_fidelity}
\end{figure}

Table~\ref{tab:latent_fidelity} first establishes the decoder reference: reconstructing the ground-truth DINOv3 feature reaches an SSIM of $0.9700$ and an FID of $7.10$ against the original future images. Conditioning the world model on the ground-truth action yields $0.9569$ SSIM and $7.40$ FID, showing that its predicted latent retains much of the target future structure after decoding. Replacing the Stage~2 proposal with the Stage~3 refined action increases SSIM from $0.9520$ to $0.9562$ and decreases FID from $7.41$ to $7.40$. The SSIM gain of $0.0042$ provides paired evidence that futures induced by refined actions align more closely with their targets. The $0.01$ FID reduction is small, and is therefore interpreted only as preserving, with a slight improvement in, set-level distributional fidelity. Stage~3 also approaches the GT-action reference, trailing it by only $0.0007$ SSIM and matching its FID at the reported precision, consistent with Figure~\ref{fig:latent_fidelity}. This diagnostic supports improved latent-future alignment after refinement, but does not by itself establish that all policy gains are caused exclusively by the imagined-future branch.

\subsubsection{Disentangling Future Feedback and Proposal Anchoring}

We next isolate the two inputs that distinguish the Stage~3 refiner from a conventional state-conditioned action head: the imagined future $\hat I_f$ and the original Stage~2 proposal $\hat a^0$. In all Stage~3 variants, the Stage~1 world model and Stage~2 proposal policy remain frozen, and we keep the refiner depth and training protocol unchanged. The full model receives all three branches,
\[
R_\psi(c_t,\hat I_f,\hat a^0).
\]
We compare it with two targeted input ablations. \textbf{Stage~3 w/o imagined future} removes the world-model rollout and trains the refiner from only the current representation and proposal,
\[
R_\psi(c_t,\hat a^0).
\]
This variant tests whether a generic state-conditioned proposal refiner can explain the Stage~3 gain without consequence feedback. \textbf{Stage~3 w/o action proposal} retains proposal-conditioned imagination but removes the proposal tokens from the refiner,
\[
\hat I_f=W_\theta(I_c,\hat a^0),
\qquad
R_\psi(c_t,\hat I_f).
\]
Thus, $\hat a^0$ is still used to generate the same imagined consequence as in the full model, but is not exposed as a correction anchor when producing the final action. This variant tests whether the current state and predicted outcome are sufficient, or whether the refiner must also know which action produced that outcome.

\begin{table}[H]
\centering
\caption{\textbf{Stage~3 input-branch ablation on LIBERO.} Current, Future, and Proposal indicate the branches provided directly to the refiner. The Stage~2 proposal policy has no refiner and serves as the unrefined reference. For Stage~3 w/o action proposal, the frozen Stage~2 proposal is still used by the world model to generate the Future branch, but is withheld from the refiner.}
\label{tab:stage3_branch_ablation}
\resizebox{\linewidth}{!}{%
\begin{tabular}{lcccccccc}
\toprule
Variant & Current & Future & Proposal & Spatial & Object & Goal & Long & Avg. \\
\midrule
$\omega$-EVA Stage~2         & $\surd$ & --      & --      & 98.8 & 99.4 & 97.6 & 95.8 & 97.9 \\
\midrule
$\omega$-EVA Stage~3                     & $\surd$ & $\surd$ & $\surd$ & 99.0 & 99.8 & 98.2 & 97.4 & 98.6 \\
\quad w/o imagined future                & $\surd$ & --      & $\surd$ & 97.6 & 99.8 & 96.8 & 94.6 & 97.2 \\
\quad w/o action proposal                & $\surd$ & $\surd$ & --      & 97.2 & 99.4 & 95.0 & 92.4 & 96.0 \\
\bottomrule
\end{tabular}%
}
\end{table}

Table~\ref{tab:stage3_branch_ablation} shows that both imagined-future feedback and proposal anchoring are necessary for the Stage~3 gain. Removing the imagined future reduces the average from $98.6\%$ to $97.2\%$, a $1.4$-point drop, and performs $0.7$ points below the unrefined Stage~2 policy. Thus, an additional state-and-proposal refiner alone does not explain the improvement; proposal-conditioned consequence feedback provides information that is absent from generic refinement. Removing the proposal branch produces a larger decline to $96.0\%$, $2.6$ points below the full model and $1.9$ points below Stage~2. Although this variant still receives the future imagined from the same Stage~2 proposal, the refiner no longer observes which action produced that consequence. The result supports the proposal's role as a correction anchor for translating future feedback into a precise action update. Together, the two ablations validate the full current--future--proposal interaction: the imagined consequence provides the feedback to assess, while the proposal identifies the action that must be rewritten.

\subsubsection{Proposal Denoising Steps and Compute--Performance Trade-off}

Stage~2 generates the proposal $\hat a^0$ by integrating a flow policy for a fixed number of Euler steps. The main model uses five steps, but this proposal is subsequently checked and corrected by the Stage~3 refiner. A preliminary diagnostic in Appendix~\ref{app:proposal_budget_diagnostic} shows that simply evaluating a refiner trained with five-step proposals using one-step proposals creates a modest training--inference mismatch, motivating matched training under the intended proposal budget. For the reduced-budget variants below, Stage~3 uses the Stage~1 world model for proposal-conditioned imagination, which the diagnostic suggests is at least as stable as the co-trained world model when proposals are only partially denoised. We therefore evaluate whether Stage~3 requires a highly denoised proposal or can preserve performance with a smaller proposal-generation budget. Each variant trains and evaluates Stage~3 with the same Stage~2 step count. We also include a zero-step random-proposal control, which replaces the denoised proposal with the initial noise sample while keeping the same refiner training protocol.

\begin{table}[H]
\centering
\caption{\textbf{Stage~2 proposal denoising steps for Stage~3 on LIBERO.} Steps denote the number of Euler integration steps used by the Stage~2 flow policy to generate the proposal. Inference time is measured per decision on a single NVIDIA RTX 4090 using \texttt{torch.compile} and bfloat16, without additional engineering optimization. Stage~3 is trained and evaluated with the same step count in each row, using the Stage~1 world model for proposal-conditioned imagination. The zero-step variant uses the initial random noise as the proposal and serves as a control for whether proposal information is still needed. Success rates are reported in percent.}
\label{tab:proposal_denoising_steps}
% \resizebox{\linewidth}{!}{%
\begin{tabular}{lcccccc}
\toprule
Steps & Time (ms) & Spatial & Object & Goal & Long & Avg. \\
\midrule
0    & 25           & 97.0 & 99.2 & 92.6 & 90.2 & 94.8 \\
1    & 25       & 99.2 & 100.0 & 99.0 & 96.8 & 98.8 \\
3    & 37           & 99.2 & 99.8 & 98.8 & 97.8 & \textbf{98.9} \\
5    & 45       & 99.4 & 100.0 & 98.4 & 95.8 & 98.4 \\
\bottomrule
\end{tabular}%
% }
\end{table}

Table~\ref{tab:proposal_denoising_steps} shows that one, three, and five proposal denoising steps yield similar average success rates, ranging from $98.4\%$ to $98.9\%$, and performance does not improve monotonically with more Stage~2 integration steps. In particular, using a single Stage~2 step reaches $98.8\%$ average success, comparable to the five-step variant at $98.4\%$, while reducing measured inference time from $45$\,ms to $25$\,ms per decision. This suggests that, when Stage~3 is trained under the intended proposal budget, the refiner can correct a coarser but still informative proposal. The random-proposal control drops to $94.8\%$, especially hurting LIBERO-Long, so the result should not be interpreted as Stage~3 ignoring the proposal branch. Instead, the ablation indicates that $\omega$-EVA can reduce the number of Stage~2 proposal iterations from five to one while maintaining comparable control performance and running at approximately $40$\,Hz on a single RTX 4090 under our measured setup. The timing result supports real-time deployment, while leaving further latency reductions from dedicated systems optimization outside the scope of this study.

%% file: chapters/5_Conclusion.tex
\section{Conclusion}

We presented $\omega$-EVA, a latent interactive world model that turns future prediction into proposal-conditioned feedback for embodied action generation. Rather than using a world model only as a training objective, representation learner, or standalone simulator, $\omega$-EVA places it inside the action-generation loop. Its three-stage framework first learns action-conditioned latent dynamics, then trains a language-conditioned flow policy to produce an initial action chunk, and finally closes the loop through Stage~3: the proposal is fed back to the world model, its latent consequence is envisioned, and a tri-branch refiner jointly reasons over the current state, imagined future, and proposal to produce the final action. Because this interaction remains in feature space, the policy can reason about a candidate consequence without generating a future video.

Experiments across LIBERO, LIBERO-PLUS, and RoboTwin~2.0 consistently show that the complete Envision--Verify--Act pipeline improves its Stage~2 proposal policy. Stage~3 raises average success from $97.9\%$ to $98.6\%$ on LIBERO, from $71.3\%$ to $72.2\%$ under zero-shot LIBERO-PLUS transfer, from $81.2\%$ to $83.4\%$ with LIBERO-PLUS training, and from $88.9\%$ to $90.3\%$ on RoboTwin~2.0. The gains cover all LIBERO suites, both RoboTwin evaluation settings, and most LIBERO-PLUS perturbation categories. The latent-fidelity analysis further provides representation-level evidence that action-conditioned predictions preserve meaningful future structure and that futures induced by refined actions align more closely with the target than those induced by Stage~2 proposals. Taken together, these results support the value of the complete interaction pipeline while leaving the independent causal contribution of each Stage~3 branch to the controlled studies now under evaluation. Notably, these results are obtained with an approximately 1.2B-parameter model and no additional robot-data pretraining, demonstrating a compact and competitive performance--scale--data trade-off rather than relying on substantially larger pretrained policies.

\paragraph{Future directions.}
The present system performs one consequence-aware refinement before executing an action chunk and replans after receiving the next observation. A natural extension is \emph{intra-chunk closed-loop refinement}, in which intermediate observations continuously update the imagined future and remaining actions while the chunk is being executed. A complementary direction is \emph{iterative imagination--refinement}: a refined action can be returned to the world model to produce a new consequence, followed by another refinement step. Studying the number of iterations, convergence behavior, latency, and task success would expose the trade-off between deeper consequence reasoning and responsive control. The predictive interface can also be extended beyond vision by incorporating tactile, force, and proprioceptive signals, which may be especially valuable under contact, occlusion, and visual ambiguity. Although the current compact model is trained only on benchmark robot data, robot-video or trajectory pretraining and larger model scales may improve dynamics fidelity and generalization. Finally, action-conditioned world modeling offers a promising interface for test-time self-evolution: after executing an action, the agent can compare its imagined consequence with the observed outcome and use this feedback to continually calibrate the world model and, potentially, adapt the policy and refiner. These directions point toward embodied world-model agents that do not merely imagine once before acting, but continuously perceive, envision, revise, and learn from their own interaction.

%% file: chapters/6_Appendix.tex
\section{Additional Ablation Diagnostics}

\subsection{World-Model Source for Reduced Proposal Budgets}
\label{app:proposal_budget_diagnostic}

Table~\ref{tab:wm_source_diagnostic} reports a compact diagnostic for the Stage~2 proposal budget ablation in Section~4. The purpose of this study is not to define the final performance setting, but to explain why the matched step-budget experiment uses the Stage~1 world model for proposal-conditioned imagination. The original Stage~3 refiner is trained with five-step Stage~2 proposals and the Stage~2 co-trained world model. Evaluating the same refiner with one-step proposals introduces a proposal-distribution shift and reduces average success from $98.6\%$ to $97.7\%$. Replacing the co-trained world model with the Stage~1 world model has little effect for clean five-step proposals, but gives a small improvement for one-step proposals. This suggests that the Stage~1 world model is at least not worse under clean proposals and may be more stable when proposals are only partially denoised, motivating the matched-budget study in Table~\ref{tab:proposal_denoising_steps}.

\begin{table}[H]
\centering
\caption{\textbf{Diagnostic for world-model source and proposal-budget mismatch on LIBERO.} The refiner is the original Stage~3 model trained with five-step proposals. Changing the inference proposal steps or the world-model source is used only as a diagnostic for distribution shift and consequence-model robustness. Success rates are reported in percent.}
\label{tab:wm_source_diagnostic}
\begin{tabular}{lccc}
\toprule
Variant & Imagination WM & Inference steps & Avg. \\
\midrule
Original five-step proposal & Stage~2 co-trained & 5 & 98.6 \\
One-step proposal mismatch & Stage~2 co-trained & 1 & 97.6 \\
Five-step proposal with Stage~1 WM & Stage~1 & 5 & 98.4 \\
One-step proposal with Stage~1 WM & Stage~1 & 1 & 98.0 \\
\bottomrule
\end{tabular}
\end{table}